\pdfoutput=1

\documentclass[11pt]{article}

\usepackage[]{acl}

\usepackage{times}
\usepackage{latexsym}

\usepackage[T1]{fontenc}

\usepackage[utf8]{inputenc}

\usepackage{microtype}

\usepackage{inconsolata}

\usepackage{enumitem}
\usepackage{subcaption}

\usepackage{xcolor}
\usepackage{graphicx}
\usepackage{tabularx}
\usepackage{soul}
\usepackage{booktabs}
\usepackage{multirow}

\definecolor{lightgreen}{RGB}{144,238,144}  
\definecolor{lightgray}{RGB}{220,220,220}

\newcommand{\ap}[1]{{\textcolor{black}{#1}}}

\newcommand{\bs}[1]{{\textcolor{black}{#1}}}

\title{A \textit{Prompt} Response \\
to the Demand for Automatic Gender-Neutral Translation}

\author{Beatrice Savoldi,\textsuperscript{1} Andrea Piergentili,\textsuperscript{1,2} 
\textbf{Dennis Fucci},\textsuperscript{1,2} \textbf{Matteo Negri}\textsuperscript{1}, \textbf{Luisa Bentivogli}\textsuperscript{1}\\
  \textsuperscript{1} Fondazione Bruno Kessler\\
  \textsuperscript{2} University of Trento\\
  {\tt \{bsavoldi,apiergentili,dfucci,negri,bentivo\}@fbk.eu} \\}

\begin{document}
\maketitle
\sethlcolor{lightgray}
\begin{abstract}
Gender-neutral translation (GNT) that avoids biased and undue binary assumptions is a pivotal challenge for the creation of more inclusive translation technologies.  Advancements for this task in Machine Translation (MT),  however, are hindered by the lack of dedicated parallel data, which are necessary to adapt MT systems to satisfy neutral constraints.
For such a scenario, large language models 
offer hitherto unforeseen possibilities, as they come with the distinct advantage of being versatile in various (sub)tasks 
when provided with explicit instructions.
In this paper, we explore this potential to automate GNT by comparing 
MT 
with the 
popular GPT-4 model. Through extensive 
manual analyses, our study empirically reveals the inherent limitations of current MT systems in generating 
GNTs
and provides valuable insights into the potential 
and challenges
associated with
prompting
for neutrality.

\end{abstract}

\section{Introduction}
\label{sec:intro}


To foster greater inclusivity in our communication practices, there has been a 
rise in the adoption of gender-neutral language strategies  \citep{hord2016bucking, Euguideline}, which challenge gender norms and embrace all identities by eschewing unnecessary gendered terms 
(e.g. \textit{police officer} vs \textit{policeman}). 
Such strategies are now widespread across various domains – including institutions \citep{hoglund2023gendering}, academia \citep{apa2020publication}, and industry \citep{microsoft}, with their consequential 
investigation for
various natural language processing (NLP) technologies \citep{cao2020toward, brandl-etal-2022-conservative, wagner-zarriess-2022-gender}. 

While recent advancements in NLP have seen the modeling of neutral language into monolingual applications \citep{vanmassenhove-etal-2021-neutral, sun2021they, amrhein-etal-2023-exploiting, veloso-etal-2023-rewriting}, research 
in cross-lingual settings is relatively limited.
%
%
%
%
%
Previous works in MT \citep[\textit{inter alia}]{costa-jussa-de-jorge-2020-fine,savoldi2021gender,choubey2021improving, alhafni-etal-2022-user, Piazzolla_Savoldi_Bentivogli_2024} have been mostly confined within binary perspectives to improve the generation of masculine/feminine forms into grammatical gender languages (e.g. \textit{doctors} $\rightarrow$ it: \textit{dottor\textbf{i}/\textbf{esse}}).\footnote{
Although in grammatical gender languages also inanimate nous are formally assigned to 
a gender class \citep{Corbett:91}, we are hereby only concerned with (social) gender assignment for human referents.}
%
%
%
%
Under realistic scenarios though, systems 
often 
encounter ambiguous input sentences that do not convey 
gender distinctions \citep{saunders-2023-improving, piergentili-etal-2023-gender}, and for which 
GNT would be
preferable to prevent undue gender 
assignments
in the target language 
(e.g. en:  \textit{\textbf{doctors} $\rightarrow$ it: \textbf{personale medico}\footnotesize{\texttt{[the medical staff]}}}).

Despite 
individual studies indicating that existing MT systems are ill-equipped to handle neutrality  \citep{cho-etal-2019measuring,piergentili-etal-2023-hi, savoldi-etal-2023-test}, 
the automation of 
GNT
remains an open challenge, hampered by the lack of dedicated 
resources. To the best of our knowledge,  the work by \citet{saunders-etal-2020-neural} 
stands as the sole effort to create gender-neutral MT models, but their fine-tuning approach does not 
generalize from their small artificial adaptation set.
%
Within this landscape, 
large
language models 
(LLMs)
can offer a solution to meet the demand for gender neutrality, thanks to their adaptability to perform new (sub)tasks based on explicit instructions 
and few examples
\citep{brown-2020-learners}. 
In fact, albeit 
LLMs
still lag slightly behind traditional MT 
in overall 
translation 
quality \citep{robinson_chatgpt_2023,vilar_prompting_2023,zhang_prompting_2023}, 
their
versatility for controlling specific 
aspects in the output translation 
was proven for several attributes \citep{moslem-etal-2023-adaptive, sarti-etal-2023-ramp, garcia_using_2022, yamada_optimizing_2023}.

In this paper, we thus 
seek to advance the automation of neutral translation by exploring the potential of instruction-following 
models.
To this aim, we focus on English$\rightarrow$Italian 
and 
systematically compare the neutral capabilities of 
traditional MT models with 
GPT-4 \citep{openai2023gpt4}. 
By experimenting with different prompts and shot-exemplars, 
we conduct a fine-grained, 
manual
evaluation showing that:
\textit{\textbf{i)}} 
used 
\textit{as is} neither MT 
nor
GPT are suitable for GNT, but prompting GPT shows surprising neutralization capabilities 
elicited with just 
a few examples;
\textbf{\textit{ii)}} while including 
test set terms as neutralization exemplars in the prompts leads to
slightly better GNT performance,
GPT can generalize well also when provided with unseen 
examples. Finally, 
extensive manual evaluations unveil that \textbf{\textit{iii)}} 
judging the  quality and acceptability of automatic 
GNT
is a 
subjective task, with notable variations across annotators.
%
%
%
%
To promote future research, 
we make all our manual output annotations freely available at: \url{https://mt.fbk.eu/gente/}.\footnote{Released under a Creative Commons Attribution 4.0 International license (CC BY 4.0).}
%
%
%


\section{Methods and Settings}
\label{sec:settings}
\paragraph{Test set.}
\label{subsec:test-set}
We run our experiments on GeNTE
\citep{piergentili-etal-2023-hi}, a recently 
released
parallel test set
designed to evaluate  models' 
GNT capabilitites.
Built on Europarl data \citep{koehn2005europarl}, it allows us to test 
MT on 
naturalistic 
instances for en-it,
a language pair that is 
highly 
representative of the challenges of performing GNT into 
languages with extensive gendered morphology.
For such languages, neutral strategies can range from simple word changes 
(e.g. omissions or synonyms)
to 
complex 
reformulations that 
can 
alter the sentence structure \citep{gabriel2018neutralising}. 
Hence, generating 
suitable GNTs is a delicate and difficult task, to be carefully weighted
not to
impact the acceptability of a translation.
Here, we use 
a 
portion of GeNTE
consisting of 750 English sentences that are gender-ambiguous,\footnote{\texttt{Set-N} in the original corpus.} 
and which are thus to be neutrally translated so as to avoid any undue gender inference in Italian (e.g. \textit{I,  with \textbf{all my colleagues} wish to...},  it-M: \textit{Io, con \textbf{tutt\underline{i} \underline{i} collegh\underline{i}} desidero...} $\rightarrow$  it-GNT: \textit{Io, con \textbf{ogni collega}\texttt{\footnotesize{[each colleague]}}, desidero...}).\footnote{For more details, see Appendix \ref{app:GeNTE}.}


%



\paragraph{Systems.}

As MT models, we select two state-of-the-art commercial systems: 
Amazon Translate\footnote{\url{https://aws.amazon.com/it/translate/}.} 
and
DeepL.%
\footnote{\url{https://www.deepl.com/en/translator}.}
For 
\textsc{GNT-prompting}, we  
use GPT (\texttt{gpt-4-0613}),
which 
achieved
%
promising results in translation 
\citep{jiao2023chatgpt},
though 
especially
for high-resource languages \citep{robinson_chatgpt_2023,stap-araabi-2023-chatgpt}.
As an \textit{instruction-following} model
\citep{chung2022scaling,ouyang_training_2022}, GPT
is suited to 
keep
adherence to provided guidance when performing a task, a valuable aspect 
to control the 
neutral translation 
of 
gendered terms.


\begin{table}[t]
\centering
\scriptsize
\begin{tabular}{l|cccc}
 \toprule
 &
  \multicolumn{1}{c}{\textbf{BLEU}} &
  \multicolumn{1}{c}{\textbf{CHRF}} &
  \multicolumn{1}{c}{\textbf{BLEURT}} &
  \multicolumn{1}{c}{\textbf{COMET}} \\
  \toprule
\textbf{Amazon}     & 31.04 & 57.54 & 82.84 & 84.07 \\
\textbf{DeepL}      & 30.75 & 56.30 & 82.80 & 83.90 \\
\textbf{GPT-4} & 25.08 & 51.94 & 80.56 & 82.60 \\
\bottomrule
\end{tabular}
\caption{Overall quality results for en-it.}
\label{tab:translation_quality}
\end{table}

\begin{table*}[t]
\scriptsize
\begin{tabularx}{\textwidth}{lX}
\toprule
\texttt{Contr} & \begin{tabular}[c]{@{}p{14.5cm}@{}} {[English]}: Secondly, how far does it increase transparency and accountability \textbf{of the writers}? \\
{[Italian, gendered]}: Secondariamente, fino a che punto aumenta la trasparenza e la responsabilità \textbf{degli scrittori}?\\
{[Italian, neutral]}: \hl{Secondariamente, fino a che punto aumenta la trasparenza e la responsabilità \textbf{di chi scrive}?}
\\\end{tabular} \\ \midrule

\texttt{CoT-src} & \begin{tabular}[c]{@{}p{14.5cm}@{}}Q: Translate the following English sentence into Italian using a gender-neutral language to refer to human entities: [Secondly, how far does it increase transparency and accountability \textbf{of the writers}]. Think step by step.
\vspace{3pt}
\\ A: In the English sentence 
there is one expression which refers to human entities and could be translated in a non-neutral way: <of the writers>. A gender-neutral translation of <of the writers> is <di chi scrive>. The final gender-neutral translation is [\hl{Secondariamente, fino a che punto aumenta la trasparenza e la responsabilità \textbf{di chi scrive}?]} \\
\end{tabular} \\ \midrule

\texttt{CoT-tgt} & \begin{tabular}[c]{@{}p{14.5cm}@{}}Q: Translate the following English sentence into Italian using a gender-neutral language to refer to human entities: [Secondly, how far does it increase transparency and accountability \textbf{of the writers}?]. Think step by step.  
\vspace{3pt}
\\
A: The English sentence 
can be translated as [Secondariamente, fino a che punto aumenta la trasparenza e la responsabilità \textbf{degli scrittori}?]. There is one «expression with <non-neutral terms>» that refers to human entities: «<degli scrittori>». A gender-neutral alternative to «<degli scrittori>» is «di chi scrive». The final gender-neutral translation is [\hl{Secondariamente, fino a che punto aumenta la trasparenza e la responsabilità \textbf{di chi scrive}?}].
\end{tabular} \\
\bottomrule

\end{tabularx}
\caption{
Examples of each prompt template. The source ``\textbf{\textit{of the writers}}'' -- corresponding to the gendered ``\textit{\textbf{degl\underline{i} scrittor\underline{i}}}'' in Italian -- is neutralized as ``\textit{\textbf{\footnotesize{di chi scrive}}}'' \texttt{[of who writes]}. \texttt{CoT-tgt} and \texttt{CoT-src} templates are structured as Questions and Answers. 
The final \hl{gender-neutral translations} are highlighted.}
%
\label{tab:prompts-in-domain}
\end{table*}

\paragraph{Experiments.}
We explore 
models' neutralization 
abilities
under two experimental settings: \textit{i)} 
\textsc{Baseline}, 
to 
compare if 
the MT 
models 
and GPT 
in zero-shot conditions\footnote{We adopt the best performing prompt by \citet{peng2023making}: \texttt{``Please provide the [TGT] translation of the following sentence:''}.}
can perform GNT, without being explicitly instructed/adapted for the task; and \textit{ii)} 
\textsc{GNT-prompting},
to leverage GPT potential  when prompted with 
dedicated
instructions and 
examples.
In both settings, 
for GPT 
we 
use temperature
$0.0$, since \citet{peng2023making} attested a progressive translation degradation 
with 
higher temperature values.

Before delving into their GNT capabilities,
in Table \ref{tab:translation_quality} we report the 
performance
of 
all
models 
on the Europarl common test set.\footnote{
\url{https://www.statmt.org/europarl/}.}
%
Such results 
confirm that GPT exhibits good cross-lingual capabilities, but 
does not match 
traditional MT models. 


\section{\textsc{GNT-prompting}}
\label{sec:gnt-prompting}
To elicit 
GPT's flexibility 
for
neutral translations, in the 
\textsc{GNT-prompting}
condition we experiment with three few-shot templates inspired by 
existing 
literature on 
prompting \citep{liu-prompt,dong_survey_2023}. 
Our prompts, shown in Table \ref{tab:prompts-in-domain}, are:


\textbf{(1)} \textbf{\texttt{Contr}}:  consisting of \textit{contrastive} examples of gendered and neutral translations for each English sentence, without 
additional verbalized instructions. 
This
simple template has shown promising results for controlling the generation of (binary) gender forms
\citep{sanchez_gender-specific_2023}. \\
\indent \textbf{(2)} \textbf{\texttt{CoT-src}}: 
based on \textit{chain-of-thought} demonstrations that break complex tasks into intermediate reasoning steps  \citep{wei_chain--thought_2023}. This 
prompt
first guides the identification of \textit{source} terms that correspond to a gendered expression in Italian, then 
elaborates
on the neutralization of each term to provide the final target translation. \\
%
%
%
\indent \textbf{(3)} \textbf{\texttt{CoT-tgt}}: 
%
similar to \texttt{CoT-src}, but with 
different 
steps, i.e. this 
prompt 
provides an (intermediate) gendered translation
and 
identifies
the \textit{target} terms  to be neutralized in the final translation.

Each 
prompt
is used with 
3
exemplar
sentences
taken from the institutional domain,
a context where neutral language is 
increasingly
employed, and which is also 
covered by
GeNTE.
To 
verify GPT's ability to generalize from the provided examples, we experiment with two sets of 
sentences, which 
only differ for the inclusion of terms to be neutralized that 
are either \textit{i)} present in GeNTE  -- hence \textit{seen} --  or \textit{ii)} terms that 
never occur in the test set -- hence \textit{not seen}. 
We   list such terms in Table \ref{tab:terms}, whereas we refer to Appendix \ref{app:prompts} for further details concerning our prompting experiments.


\begin{table}[t]
    \centering
    \scriptsize
    \begin{tabular}{cc|cc}
    \toprule
    \multicolumn{2}{c|}{\textbf{Seen}} & \multicolumn{2}{c}{\textbf{Not seen}} \\
    \midrule
    \textbf{en} & \textbf{it} & \textbf{en} & \textbf{it} \\
    \midrule
        MEPs & parlamentari europei  & writers & scrittori \\
        President & Signora Presidente & manager & direttore \\
        everyone & tutti & employees & impiegati \\
        politicians & politici & musicians & musicisti \\
        \multirow{2}{*}{fishermen} &   \multirow{2}{*}{pescatori}
        & \multirow{2}{*}{freshmen} & \multirow{2}{*}{\shortstack{studenti \\ del primo anno}} \\
       
    \end{tabular}
    \caption{Source English and target Italian pairs of \textit{seen} and \textit{not seen} terms used in the exemplar sentences.}
    \label{tab:terms}
\end{table}

\begin{figure*}[t]
 \begin{subfigure}[b]{0.26\textwidth}
    \includegraphics[scale=0.30]{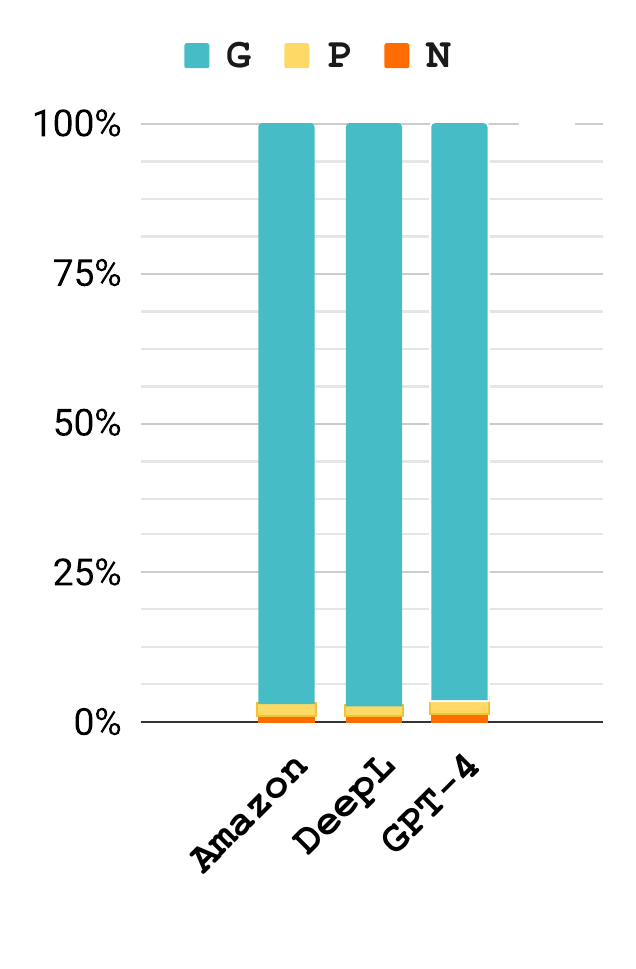}
    \caption{\textsc{Baseline} neutrality.}
     \label{fig:zero-shot-results}
  \end{subfigure}
  \begin{subfigure}[b]{0.32\textwidth}
    \includegraphics[scale=0.30]{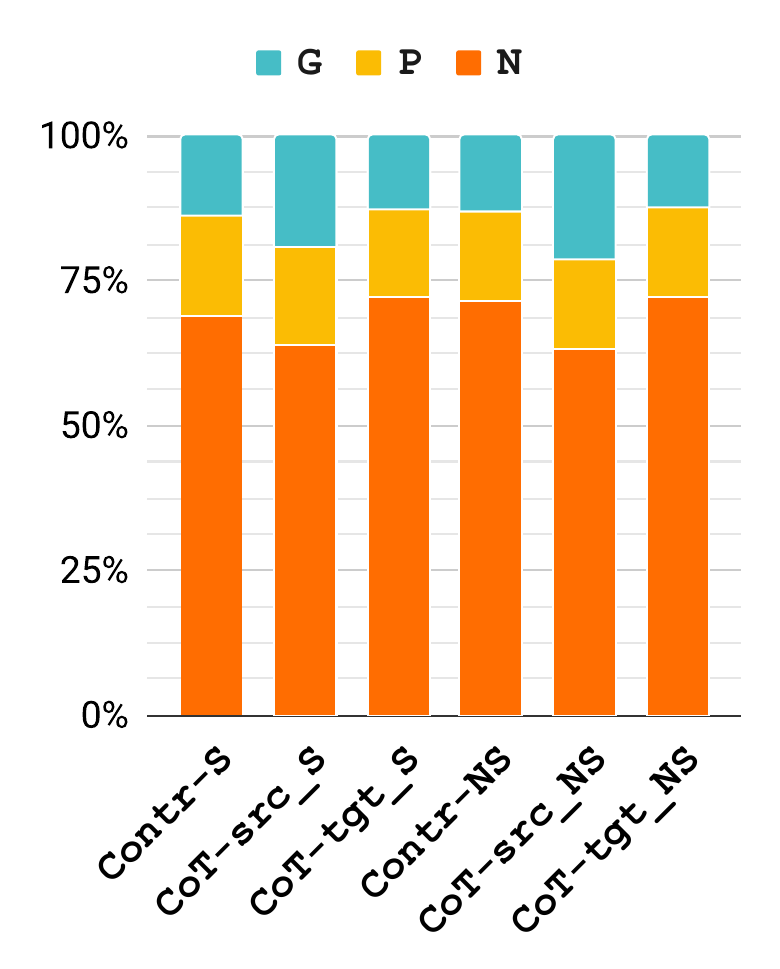}
    \caption{\textsc{GNT-Prompting} neutrality.}
        \label{fig:gnt-results}
  \end{subfigure}
  \begin{subfigure}[b]{0.3\textwidth}
    \includegraphics[scale=0.30]{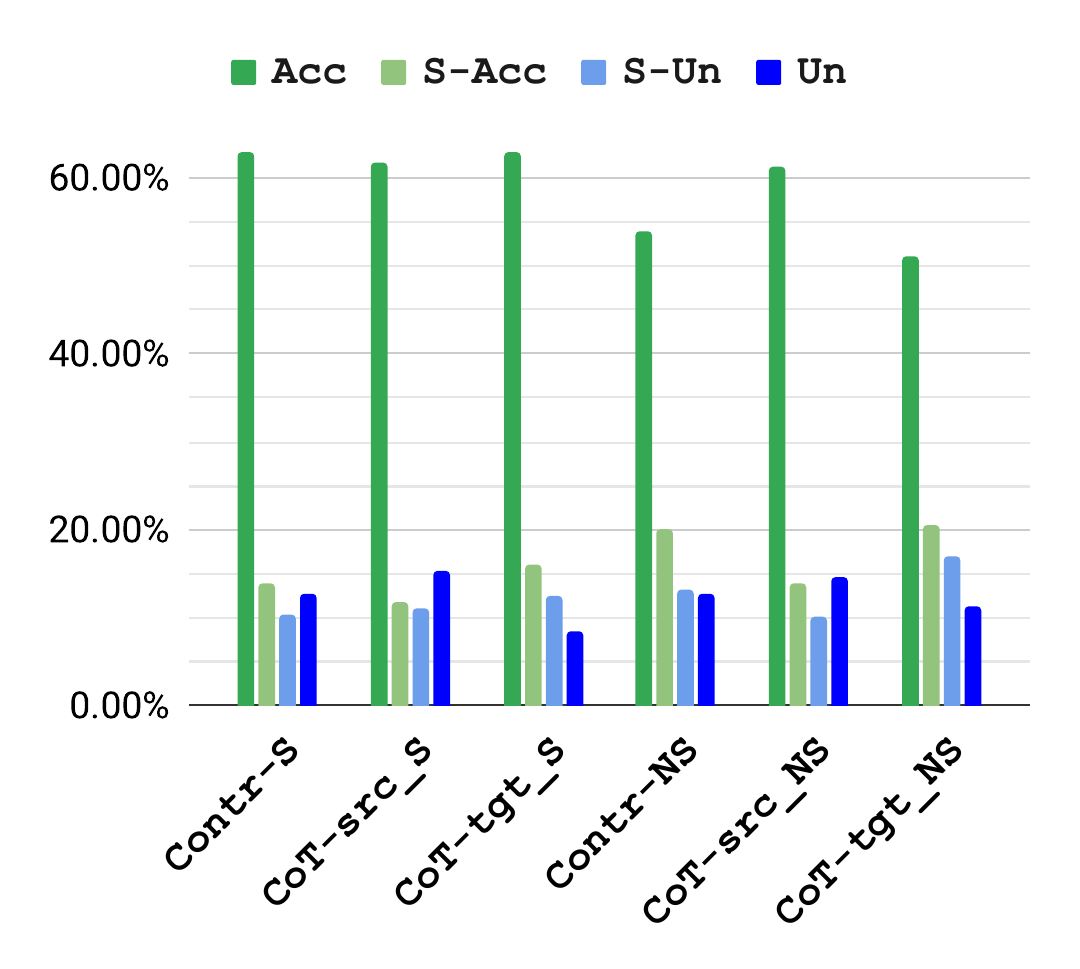}
    \caption{\textsc{GNT-prompting} acceptability.}
     \label{fig:acc-results}
  \end{subfigure}
  \caption{Manual Evaluation Results.\protect\footnotemark{}}
\end{figure*}

\section{Manual Evaluation Results}
\label{sec:manual-eval-res}
In this section, we present the results obtained by all our models in \textsc{Baseline} conditions, and by GPT in 
\textsc{GNT-prompting}
conditions.
Although the assesment of GNT capabilites can be automated with the official GeNTE evaluation protocol,
the approach would present
two inherent limitations. Since the protocol simply classifies 
whether
the whole output translation is gendered or neutral, it does not  %
consider
neutralization success/failure for 
multiple
terms in the sentence individually, nor the correctness and acceptability of the 
corresponding
translations.\footnote{E.g., \textit{I am \textbf{happy}} $\rightarrow$ \textit{Sono \textbf{triste}} (``sad'') counts as a -- implicitly correct -- neutralization, despite its inadequacy.} 
%
To 
account for these
aspects, we 
hence resort to a
\textbf{two-layered} 
\textbf{manual evaluation} that first 
%
%
%
%
distinguishes \textit{i)}  fully Neutral (\texttt{N}) and  \textit{ii)} fully Gendered (\texttt{G}), from \textit{iii)} Partially neutral
(\textsc{\texttt{P}}) outputs where one or more gendered expressions in the sentence are not neutralized.
%
%
%
Then, we judge 
whether
the generated GNTs are 
acceptable
(i.e. if they sound fluent and adequately represent the source meaning)
on the Likert scale
\textit{i)} acceptable (\texttt{Acc}), \textit{ii)} somewhat acceptable (\texttt{S-Acc}), \textit{iii)} somewhat unacceptable (\texttt{S-Un}), 
\textit{iv)} unacceptable (\texttt{Un}).\footnote{More information on the manual analysis setup and guidelines is provided in 
Appendix \ref{app:analyses}.} Example 
judgements are 
shown
in Table \ref{tab:annotation_examples}.

\begin{table}[t]
\setlength{\tabcolsep}{4pt}
\scriptsize
\begin{tabular}{cllcc}
\toprule
& & \textbf{Examples} & 
\textbf{Neut.} 
& \textbf{Acc.}   \\
 \midrule
A& \textsc{Src} & I am \textbf{pleased} to make my contribution. & \multirow{2}{*}{\texttt{G}} & \multirow{2}{*}{--}   \\
& \textsc{Out} & Sono \textbf{\textit{liet\underline{o}}} di potere contribuire. &  &   \\
\midrule
B & \textsc{Src} & Respect for standards lies with \textbf{the judges}. & \multirow{2}{*}{\texttt{N}} & \multirow{2}{*}{\texttt{Acc}}   \\
& \textsc{Out} & ... spetta \textbf{\textit{all'autorità giudiziaria}}. & & \\ 
& & \texttt{[judicial authority]} & & \\
%
\midrule
C & \textsc{Src} & May I quote three \textbf{actors} in this field. & \multirow{2}{*}{\texttt{N}} & \multirow{2}{*}{\texttt{Un}}  \\
& \textsc{Out} & \begin{tabular}[c]{@{}l@{}}Posso citare tre \textit{\textbf{persone}} \texttt{{[}people{]}}...
\end{tabular} &  &    \\
\midrule
D & \textsc{Src} & \begin{tabular}[c]{@{}l@{}}\textbf{Commissioner}, I would like to \\ congratulate \textbf{the rapporteur}.\end{tabular} & \multirow{2}{*}{\begin{tabular}[c]{@{}c@{}}\texttt{P}\\ \end{tabular}} & \multirow{2}{*}{\begin{tabular}[c]{@{}c@{}}\texttt{S-Acc}\end{tabular}}   \\
& \textsc{Out} & \textit{\textbf{Commissari\underline{o}}}, vorrei congratularmi & & \\
& & con \textit{\textbf{chi ha redatto la relazione}}. & & \\
 & & \texttt{{[}who wrote the report{]}} & & \\
 
\bottomrule
\end{tabular}
\caption{Output examples with annotations.} 
\label{tab:annotation_examples}
\end{table}

For each model and prompt, 
we analyze 
the same 
200 randomly selected and anonymized output sentences,
equally distributed across three evaluators --  
all Italian native speakers, highly familiar with neutral language.\footnote{They are authors of the paper.} While each annotator worked independently, for each system we ensured a 10\% of output sentences judged by all raters to verify inter-annotator agreement (IAA). 

For the first annotation layer (\texttt{G,N,P}), 
the Fleiss' kappa
on label assignment \citep{fleiss1971measuring} amounts to 0.89, which corresponds to ``almost perfect agreement'' \citep{landis}. Disagreements were all oversights and
thus reconciled.

For the acceptability annotations, instead, we 
measure IAA with the intraclass 
correlation coefficient (ICC)\footnote{We use the statistical analysis package Pingouin to compute the \texttt{ICC3} score: \url{https://pingouin-stats.org/build/html/generated/pingouin.intraclass_corr.html}.}
\citep{fisher1925statistical, shrout1979icc}. 
In this way, rather than solely focusing on 
label assignments (i.e. \texttt{Acc, S-Acc, S-Un, Un)} we can account for the actual distance in scores across raters on the 4-point acceptability Likert scale, and thus capture when annotators strongly disagree (e.g. \texttt{Acc} vs. \texttt{Un}) with respect to closer judgements (e.g. \texttt{Acc} vs. \texttt{S-Acc}).
%
%
%
The resulting ICC amounts to 0.48. Thus, and as we further discuss in section \S\ref{sec:gnt-results}, judging acceptability emerges as a more complex and variable task featuring 
moderate agreement. Notably, 
the generative nature of the GNT task
does not entail a  definitive `correct' answer, and the diverse  perspectives can contribute to a range of valid judgments \citep{popovic-2021-agree, plank-2022-problem}. To acknowledge such a variability, we did not enforce reconciliation for disagreements.

\subsection{\textsc{Baseline} Results}
\label{sec:preliminary-results}

%
In Figure \ref{fig:zero-shot-results}, the results achieved by Amazon, DeepL and GPT
in the 
\textsc{Baseline}  condition empirically confirm that, \textbf{used \textit{as is}, these models are unsuitable for GNT}. They indeed generate only a discouraging   \textasciitilde3\% of neutral translations (both \texttt{N} and \texttt{P}), with a \textasciitilde97\% of the outputs comprising 
only (mostly masculine) gendered terms.
%
%
%
%
Based on qualitative insights, such sporadic neutralizations largely correspond to 
(highly probable) literal
translations, which 
incidentally avoid
gendered expressions
(e.g. src: \textit{we have \textbf{addressed}}, ref-it: \textit{ci siamo \textbf{occupat\underline{i}} \footnotesize{\texttt{[took care]}}} 
$\rightarrow$ out-it: \textit{\textbf{abbiamo affrontato}} \texttt{\footnotesize{[have addressed]}}).
%
%
%
%
The few 
neutralizations
were 
unsurprisingly
considered acceptable by all evaluators, but
their negligible amount and sporadic occurrence
 motivate testing GPT's versatility with dedicated 
prompts. 


\footnotetext{For automatic evaluation results, see Appendix \ref{app:evaluation}.}

\subsection{\textsc{GNT-prompting} Results}
\label{sec:gnt-results}

Starting from 
the distribution of generated neutralizations, Figure \ref{fig:gnt-results} provides the results achieved by GPT 
\textit{i)} for
each prompt template, and \textit{ii)} across the two sets of 
in-domain exemplars, respectively including 
gendered terms that occur in GeNTE (\texttt{S}, for \textit{seen}) and terms that are not present in the test set  (\texttt{NS}, for \textit{not seen}), for a total of six configurations (\S\ref{sec:gnt-prompting}).
A bird's eye view of these scores reveals very promising results. \textbf{Across all configurations, GPT produces a 
notable amount of GNTs} (\textasciitilde65-70\% \texttt{N} and \textasciitilde15\% \texttt{P}).
Interestingly, despite slightly lower GNT performance for \texttt{CoT-src},\footnote{We hypothesize that the lack of a contrastive gendered translation in the prompt negatively impacts the GNT task.} we do not find notable differences across templates for 
\texttt{S} and \texttt{NS}
examples, thus attesting GPT abilities to generalize to newly encountered gendered terms. 

By turning to the results in Figure \ref{fig:acc-results},\footnote{For the 10\% commonly annotated outputs, we include acceptability results by averaging the  scores provided by the three evaluators.} instead, the use of 
\texttt{NS}
exemplars 
seems
to slightly reduce the acceptability degree of the generated GNTs. Still, the results are 
overall positive, with \textbf{the best 
configurations} that \textbf{produce
over 60\% of good quality 
neutralizations,} like the one in
example B in Table \ref{tab:annotation_examples}, which ensures neutrality while fully preserving fluency and 
adequate source meaning. 
%
%
%
%
%
%
Notably, we 
attest a considerable number of somewhat acceptable (\texttt{S-Acc}) / unacceptable (\texttt{S-Un}) GNTs. Indeed, for several instances the raters found that GNT was complex to perform without compromising fluency, up to the point 
where
in \textasciitilde20-30\% of the cases the neutral rephrasings generated by GPT were considered as borderline or not completely satisfactory -- as in Table \ref{tab:annotation_examples} example D, where a ``\textit{rapporteur}'' is the person in charge of 
reporting, but not necessarily the one writing a report.
%

Indeed, the 
difficulty
of judging GNTs is also reflected in the modest 
IAA
measured for acceptability (\S\ref{sec:manual-eval-res}). Examples such as the following one attest to the complexities of determining what makes a good \ap{–} or \textit{acceptable} \ap{–} neutralization:
%
%
\begin{itemize}[itemsep=0.ex, leftmargin=*, labelindent=1.7em]
    \item[\texttt{src}:] Paramilitary groups have stepped up the murders \textbf{journalists} and human rights \textbf{activists}...
    \item[\texttt{out}:] I gruppi paramilitari hanno intensificato gli omicidi di \textbf{persone che lavorano nel giornalismo}\texttt{\footnotesize{[people working in journalism]}} e \textbf{persone attive nella difesa dei diritti umani}\texttt{\footnotesize{[people active in human right defence]}}
\end{itemize}

\noindent Two raters 
judged the GNT as \textsc{S-Acc} and \textsc{S-Un} due to the allegedly awkward repetition of ``\textit{people}''.
Instead, the third evaluator considered the GNT unacceptable due also to adequacy issues (i.e. \textit{working in journalism} does not necessarily imply to be a \textit{journalist}).  
Overall, we thus recognize different sensitivities 
with respect to the potential trade-off between adequacy, fluency and the satisfaction of neutral constraints. As such,
%
the qualitative evaluation of \textbf{GNT emerges as a subjective task}, 
even across annotators with comparable  expertise in  neutral language.
%
%
%
%
%
%
This 
holds implications not only from an evaluation perspective, but 
also
for an effective 
modeling of future
automatic GNT that accounts for such a variability \citep{kanclerz-etal-2022-ground, frenda-etal-2023-epic}.



\section{Conclusions}

In response to the rising demand for inclusive language (technologies), this 
study
has focused on the possibilities of automating the generation of gender-neutral translations. In particular, given the limitations of general-purpose MT models due to the need for dedicated parallel data, we 
have explored
the potential of GPT to produce 
gender-neutral outputs 
when translating from English into Italian. 
Through extensive, fine-grained manual analyses, we demonstrated that GPT offers promising 
avenues, as it can grapple with this complex task 
when given only a
few examples and still generalizes beyond them. Importantly, our evaluations also show that determining the acceptability of what constitutes a good, acceptable neutral translation comes with notable subjectivity. To enable future research, all  
our manual output annotations
are made available
\footnote{\url{https://mt.fbk.eu/gente/}.}
to the community
to explore the modeling and 
assessment of such variability. 







\section{Limitations}

Naturally, this work comes with several limitations.

\textbf{One language pair.} Our experiments are carried out for en-it only, and we are thus cautious to indiscriminately generalize our findings. 
Nonetheless, Italian
is a highly representative example of the challenges faced in cross-lingual transfer from English. Accordingly, we believe that our observations
can broadly apply to other target grammatical gender languages for high-resource scenarios, too. 
Crucially, the decision to work on en-it was determined by the fact that -- to the best of our knowledge -- the bilingual GeNTE corpus (\S\ref{sec:settings}) is the only available resource for testing GNT.

\textbf{Closed-source models.} The study relies on different closed-source models.
This has reproducibility consequences, since these systems are regularly updated, thus potentially yielding future results that differ from those reported in this paper. 
As a first attempt to a new, complex task with relevant societal impact such as GNT, we considered reasonable to \textit{i)} focus on general-purpose models used 
at
scale by millions of users 
\textit{ii)} experiment GNT prompting on the strong GPT model, which as of October 2023 holds the first position on the AlpacaEval leaderboard.\footnote{\url{https://tatsu-lab.github.io/alpaca_eval/}.}
In the future, we plan to test open-source models for this task
and investigate  how to 
weigh
the strengths of MT (i.e. higher translation quality) with those of LLMs (i.e. adaptability to neutral constraints).

\textbf{Prompts configurations.}
%
We tested the use 
gender terms occurring/not occurring in GeNTE for
prompt 
exemplar sentences (\S\ref{sec:gnt-prompting}), so as to investigate GPT's ability to generalize from the given examples. We recognize that a more comprehensive investigation of GPT's generalization ability would advocate for the use of 
sentence exemplars 
from varying domains, with more radical structural and stylistic differences. However, for this first exploration we followed existing studies 
advocating for the choice of demonstrations based on input stylistic and semantic similarity \citep{zhang_prompting_2023,vilar_prompting_2023,agrawal_-context_2023}.

\textbf{Evaluation.}
By relying on manual analyses (\S\ref{sec:manual-eval-res}), 
we enabled
a comprehensive GNT evaluation, and overcame the shortcomings of available automated protocols. To provide an alternative method was beyond the scope of this paper, though.   
Also, although we attest moderate agreement for the GNT acceptability judgments, it should not be regarded as
a shortcoming of our evaluation procedure. Rather, on the one hand, it highlights the nuances of judging open-ended generations, for which multiple solutions and subjective perspective are valid \citep{basile-etal-2021-need, rottger-etal-2022-two}. On the other, as newly emerging forms, the perceived acceptability of neutral language is highly 
dependent
on people's attitudes and exposure to such forms, and it is reasonable to 
expect
that they will change over time \citep{Koeser}. Among other aspects, our annotated sentences could also allow to \textit{i)} model this subjectivity, and \textit{ii)} track the acceptability trajectory of GNT in time.

\section{Ethics Statement}

By investigating 
the automation of gender-neutral translation, 
this work has an inherent ethical component. In particular, it is concerned with the impact of translation technologies that
reflect exclusionary language, which potentially reinforces stereotypes, masculine 
visibility, and preclude the representation of non-binary gender identities.\footnote{We use non-binary as an umbrella term to encompass all 
identities 
within and outside the
masculine/feminine binary, and that are not represented by binary language expressions. }
Specifically, here 
we focus on 
gender-neutralization techniques that rework existing forms and grammars to avoid using needless gendered terminology, and which are endorsed by several institutions (e.g. universities, the EU).  These tactics can be viewed as an example of Indirect Non-binary Language (INL) \citep{artemistrans}, which 
prevent misgendering by eschewing gender assumptions and, as we do in this paper, \textit{equally elicit} all gender identities in language \citep{strengers-2020}. Instead, to  \textit{enhance} the visibility of non-binary individuals, Direct Non-binary Language \citep{artemistrans} 
resorts to the creation of neologisms, neopronouns, or even neomorphemes \citep{lauscher-etal-2022-welcome}.
%
Therefore,
many concurring forms can fulfill the demand for inclusive language \citep{Comandini_2021, knisely2020franccais, lardelli2023genderfair}. It is thus important to emphasize that the neutralizing techniques implemented in our work are 
not prescriptively intended.
Instead, they are orthogonal to other approaches and non-binary expressions for inclusive language (technologies)  \cite{lauscher-etal-2023-em,rivas-ginel-2022-all-inclusive}.

\section*{Acknowledgements}

This work is part of the project ``Bias Mitigation and Gender Neutralization Techniques for Automatic Translation'', which is financially supported by an Amazon Research Award AWS AI grant.
Moreover, we acknowledge the support of the PNRR project FAIR - Future AI Research (PE00000013),  under the NRRP MUR program funded by the NextGenerationEU.

\bibliography{anthology,custom}

\appendix


\section{Test set and GNT}
\label{app:GeNTE}

The GeNTE corpus \citep{piergentili-etal-2023-hi} represents, to the best of our knowledge, the only available resource for neutral translation into grammatical gender languages and for a variety of gender phenomena. 
The only other resource being the synthetic dataset by \citet{cho-etal-2019measuring}, which only focuses preserving \textit{pronouns} neutrality for
English$\rightarrow$Korean, namely into a genderless target language \citep{stahlberg2007representation}. The dataset  INES \citep{savoldi-etal-2023-test}, instead, focuses on inclusive translation from a grammatical gender language -- namely German -- into English. 

For each of its entry sentences, GeNTE includes aligned \textit{i)} source English, \textit{ii)} gendered reference translation, and \textit{iii)} gender-neutral references translation triplets. 
The 750 sentences which we are focusing on contain at least one -- and potentially several more -- source expressions corresponding to Italian gendered terms that require to be either neutralized. 
Their gendered translations corresponds to the original Europarl references \citep{koehn2005europarl}, which propagate the use of masculine generics to refer to generic referents (e.g., en: \textit{It represents a threat to \textbf{man} and animals}$\rightarrow$ ref-g: \textit{Rappresenta una minaccia per \textbf{l'uomo} e gli animali}) or assign target masculine forms to unspecified referents (e.g., en: \textit{\textbf{All the citizens}}$\rightarrow$ ref-g: \textit{\textbf{Tutti i cittadini}}). 
The neutral translations are created by replacing the gendered expressions and terms with neutral alternatives (e.g. \textbf{\textit{essere umano}}\texttt{\footnotesize[human beings]}, \textbf{\textit{tutta la cittadinanza}}]\texttt{\footnotesize{[the whole citizenship]}}) 
with different degrees of interventions to ensure \textit{i)} adherence to the source meaning, and \textit{ii)} fluency in the target language, so to avoid perceiving the use of neutral language as intrusive and unsuitable. Accordingly, for each source gender-ambiguous human entity it is ensured that a gender-neutral translation in the target language is feasible.

\begin{table*}[t]
    \centering
    \footnotesize
    \begin{tabular}{ll}
    \toprule
    \multicolumn{2}{c}{\textbf{Seen}} \\
    \midrule
    
    \texttt{SRC} & \begin{tabular}[c]{@{}p{14cm}@{}}Secondly, how far does it increase transparency and accountability \textbf{of the MEPs}?\end{tabular} \\
    \texttt{GEND} & \begin{tabular}[c]{@{}p{14cm}@{}}Secondariamente, fino a che punto aumenta la trasparenza e la responsabilità \textbf{dei parlamentari europei}?\end{tabular} \\
    \texttt{NEUT} & \begin{tabular}[c]{@{}p{14cm}@{}}Secondariamente, fino a che punto aumenta la trasparenza e la responsabilità \textbf{dei membri del Parlamento Europeo} \texttt{[of the members of the European Parliament]}?\end{tabular} \\
    \midrule
    
    \texttt{SRC} & \begin{tabular}[c]{@{}p{14cm}@{}}\textbf{President}, \textbf{everyone} must continue to adopt an ambitious approach on these issues. \end{tabular} \\
    \texttt{GEND} & \begin{tabular}[c]{@{}p{14cm}@{}}\textbf{Signora Presidente}, su tali questioni sarà necessario che \textbf{tutti} continuino a dare prova d'ambizione.\end{tabular} \\
    \texttt{NEUT} & \begin{tabular}[c]{@{}p{14cm}@{}}\textbf{Presidente} \texttt{[President]}, su tali questioni sarà necessario che \textbf{ogni persona} \texttt{[every person]} continui a dare prova d'ambizione.\end{tabular} \\
    \midrule
    
    \texttt{SRC} & \begin{tabular}[c]{@{}p{14cm}@{}}\textbf{Several fishermen} have \textbf{joined} with \textbf{the politicians} in Belgrade. \end{tabular} \\
    \texttt{GEND} & \begin{tabular}[c]{@{}p{14cm}@{}}A Belgrado, \textbf{molti pescatori} si sono \textbf{schierati} dalla parte \textbf{dei politici}.\end{tabular} \\
    \texttt{NEUT} & \begin{tabular}[c]{@{}p{14cm}@{}}A Belgrado, \textbf{molte persone che lavorano nella pesca} \texttt{[many people who work in fishery]} hanno \textbf{preso le parti} \texttt{[have taken the side of]} di \textbf{chi fa politica} \texttt{[of those who engage in politics]}.\end{tabular} \\
    \midrule
    
    \multicolumn{2}{c}{\textbf{Not seen}} \\
    \midrule

    \texttt{SRC} & \begin{tabular}[c]{@{}p{14cm}@{}}Secondly, how far does it increase transparency and accountability \textbf{of the writers}?\end{tabular} \\
    \texttt{GEND} & \begin{tabular}[c]{@{}p{14cm}@{}}Secondariamente, fino a che punto aumenta la trasparenza e la responsabilità \textbf{degli scrittori}?\end{tabular} \\
    \texttt{NEUT} & \begin{tabular}[c]{@{}p{14cm}@{}}Secondariamente, fino a che punto aumenta la trasparenza e la responsabilità \textbf{di chi scrive} \texttt{[of those who write]}?\end{tabular} \\
    \midrule
    
    \texttt{SRC} & \begin{tabular}[c]{@{}p{14cm}@{}}\textbf{HR manager}, \textbf{the employees} must continue to adopt an ambitious approach on these issues. \end{tabular} \\
    \texttt{GEND} & \begin{tabular}[c]{@{}p{14cm}@{}}\textbf{Direttore delle risorse umane}, su tali questioni sarà necessario che \textbf{gli impiegati} continuino a dare prova d'ambizione.\end{tabular} \\
    \texttt{NEUT} & \begin{tabular}[c]{@{}p{14cm}@{}}\textbf{Responsabile delle risorse umane} \texttt{[HR manager]}, su tali questioni sarà necessario che \textbf{il personale} \texttt{[the staff]} continui a dare prova d'ambizione.\end{tabular} \\
    \midrule
    
    \texttt{SRC} & \begin{tabular}[c]{@{}p{14cm}@{}}\textbf{Several freshmen} have \textbf{joined} with \textbf{the musicians} in Belgrade. \end{tabular} \\
    \texttt{GEND} & \begin{tabular}[c]{@{}p{14cm}@{}}A Belgrado, \textbf{molti studenti del primo anno} si sono \textbf{schierati} dalla parte \textbf{dei musicisti}.\end{tabular} \\
    \texttt{NEUT} & \begin{tabular}[c]{@{}p{14cm}@{}}A Belgrado, \textbf{molte matricole} \texttt{[many first-years]} hanno \textbf{preso le parti} \texttt{[have taken the side of]} \textbf{delle persone del mondo della musica} \texttt{[of the people in the music business]}.\end{tabular} \\
    
    \bottomrule
    \end{tabular}
    \caption{All the <source sentence, gendered translations, and neutral translations> triplets used as demonstrations in both the \texttt{S} and \texttt{NS} sets of examples. Relevant terms for the gendered/neutral comparison are in bold. \texttt{GNT glosses} are available in square brackets.} 
    \label{tab:all_shots}
\end{table*}

\section{Prompts}
\label{app:prompts}

This section discusses relevant aspects of the prompts used in the experiments and the interaction with GPT-4.

\paragraph{Language.} As English emerged as the most effective language for prompting \citep{shi_language_2022,zhang_prompting_2023}, we use English instructions in our prompts, except for the Italian examples in the task demonstrations.

\paragraph{Task demonstrations.}

We use 3-shots prompts, which were shown to be 
a valid compromise between performance and prompt length (i.e. affecting 
costs and inference time) in our preliminary experiments. 
%
%
The creation of 
sentence exemplars proceeded as follows:
\begin{itemize}
    \item The three initial parallel source sentences and the gendered references used in the demonstrations were selected from Europarl's en-it test set, excluding any entry that was already included in GeNTE. 
\item Source and reference translations were then  modified to the include pre-selected \textit{seen} gendered terms, which occur more than 20 times in the used GeNTE subset, and \textit{ii)} the
 \textit{unseen} terms, which never occur in the used GeNTE subset. 
 \item For such parallel sentences, all gender-neutral translations were produced by one of the evaluators, a linguist experienced with neutral language strategies. 
 \item Finally, the resulting 6 exemplar sentences (shown in Table \ref{tab:all_shots}) and their GNTs were approved by all evaluators before proceeding with the experiments.  
 \end{itemize}

\begin{table}[htp]
    \centering
    \small
    \begin{tabular}{lc}
    \toprule
        \textbf{Prompt} & \textbf{Tokens} \\
        \midrule
         \texttt{Contr\_S} & 294\\
         \texttt{Contr\_NS} & 304 \\
         \texttt{CoT-src\_S} & 560 \\
         \texttt{CoT-src\_NS}& 568 \\
         \texttt{CoT-tgt\_S}& 743 \\
         \texttt{CoT-tgt\_NS} & 781 \\
    \bottomrule
    \end{tabular}
    \caption{Number of tokens of for each of the six prompt configurations.}
    \label{tab:tokens}
\end{table}

\paragraph{Length.}
Table \ref{tab:tokens} reports the length of each prompt configuration (each template and set of sentence demonstrations) measured per number of tokens. 
The values were calculated via OpenAI's tokenizer.\footnote{\url{https://platform.openai.com/tokenizer.}}

\paragraph{Model interaction.}
We interacted with GPT-4 via the chat completions API. Iterating over the test set, we included the complete content of the prompt and the input source sentence in a single message with the \texttt{user} role. The overall cost for the generation of 200 completions for each of the three prompts with both sets of shots was 29.15\$.

\paragraph{Post-processing}
To perfom our manual analysis, we post-process GPT's output so to only extract the final neutral translations to be evaluated.

\section{Manual Analysis}
\label{app:analyses}

In our analysis, we evaluate the same set of 200 output translations for each models in the \textsc{Baseline} condition (Amazon, DeepL, GPT) and for each of the six  \textsc{GNT-prompting} configurations of GPT (i.e. \texttt{Contr/CoT-tgt/CoT-src}, with both \texttt{S} and \texttt{NS} exemplares). Hence, for a total of 9 generations and 1,800 output sentences. 
The evaluations were carried based on detailed \textbf{guidelines} -- created by the same evaluator that designed the prompt examples -- which are available with the annotated data release.


\paragraph{Evaluation Design.}
To annotate the neutrality and acceptability of the outputs sentence, we provided all evaluators with the GeNTE \textit{i)} source English sentences, and the \textit{ii)} gendered reference translations, so to allow them to -- respectively --  identify which gendered terms had to be neutralized in the output as well as judge the adequacy of the translation with respect to the input sentence. 
By design, the annotators were tasked to only focus on and judge the portions of the sentence that had to be neutralized, thus disregarding the overall quality of rest of the sentence.\footnote{To facilitate this task, we \textit{i)} automatically extracted all gendered terms in the Italian references, i.e. only words differing between the gendered and neutral reference in GeNTE, and \textit{ii)} marked them in the sentences provided to the annotators.}   
To ensure consistency and train the evaluators, 
we conducted 
a first round of trial annotations, which allowed to us to address liminal instances and identify blindspots. We have updated the final annotations guidelines accordingly.

\begin{figure}[t]
  \centering
  \includegraphics[scale=0.40]{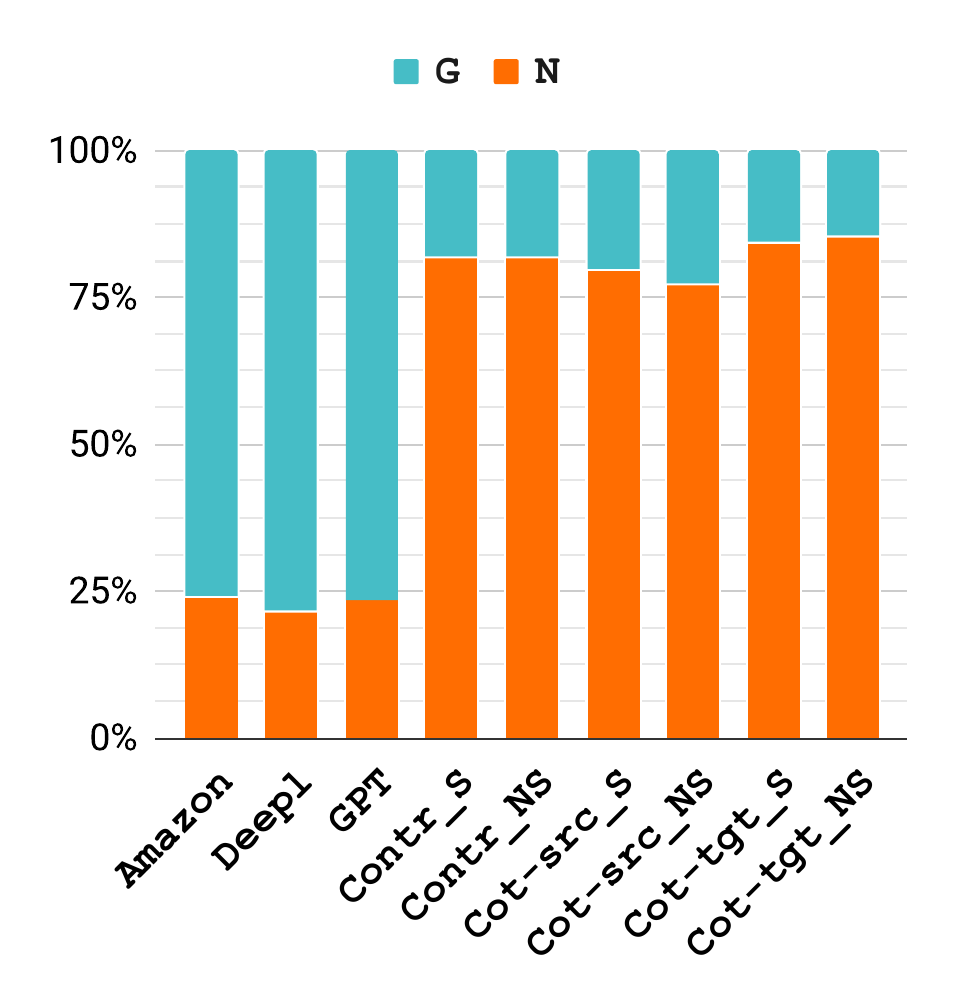}
  \caption{Neutrality for the \textsc{Baseline} and the \textsc{GNT-Prompting} settings evaluated by the classifier.}
  \label{fig:zeroshot-results}
\end{figure}

\begin{table}[t]
\centering
\small
\begin{tabular}{l||c|c|c}
\hline
\textbf{}   & \textbf{Overall} & \textbf{Neutral} & \textbf{Gendered} \\ \hline 
\hline
\texttt{Amazon}      &     85.35      & 7.84            & 86.53             \\
\texttt{DeepL}       &     86.94       & 8.70            & 88.14             \\
\texttt{GPT-4}         &     86.30        & 12.00            & 87.43             \\
\texttt{Contr\_NS}   &      74.65       & 84.69            & 49.46             \\
\texttt{Contr\_S}   &     79.30        & 87.42            & 61.22             \\
\texttt{CoT-src\_NS} &     77.55    &         85.11          &       64.41 \\
\texttt{CoT-src\_S} &    79.34         & 86.81            & 66.07             \\
\texttt{CoT-tgt\_NS} &    75.50         & 87.08            & 47.62             \\
\texttt{CoT-tgt\_S} &       79.07     & 87.90            & 55.81            
\end{tabular}
\caption{
Percentage agreement (F1 scores) between classifier-based and manual annotation evaluations, with percentages presented for both the overall agreement (weighted F1) and individual class agreements.}
\label{tab:classifier}
\end{table}

\section{Automatic Evaluation}
\label{app:evaluation}

We report the automatic evaluations results for all models and GPT configurations using the GeNTE evaluation protocol.\footnote{Classifier v2.0: \url{https://github.com/hlt-mt/fbk-NEUTR-evAL/blob/main/solutions/GeNTE.md}.}
%
As displayed in Figure \ref{fig:zeroshot-results},
the classifier's scores 
contrast with the outcomes of our manual analysis. For example, there is a 
\bs{visible}
disparity in the number of output sentences of the MT systems automatically classified as
GNTs. 
%
For this reason we exploit our manual analysis contribution to
verify the reliability of such an evaluation by calculating 
\textit{i)} Kendall's Tau
($\tau$) on the
GNT system rankings resulting from the classifier and manual analysis,\footnote{Calculated with SciPy  (\url{https://scipy.org/}).} 
and \textit{ii)} percentage agreement
calculated as F1 scores of the classifier on the
ground truth labels obtained with the manual evaluation
(see Table \ref{tab:classifier}).
To ensure a fair 
assessment of the
classifier -- which offers a binary classification (Neutral vs Gendered) -- we combined the \texttt{G} and \texttt{P} human labels. 
The $\tau$ 
coefficient yields a positive value of 0.91,  indicating that the classifier correlates very well with humans in raking systems based on the amount of generated GNTs. 
%
In general, the 
F1 results vary depending on the system, showing varying levels of satisfaction. 
F1 scores range from 7.84 for \texttt{Amazon}, where the number of true neutral sentences is notably low (as reflected in the weighted global scores), to 87.90 in the \texttt{CoT-tgt\_S} for the neutral class.
%
This calls for future investigation on the performance of the classifier, 
which is however beyond the scope of this paper.

\end{document}